\documentclass[a4paper,10pt]{amsart}

\usepackage{amsthm}
\usepackage{amssymb,amsmath}
\usepackage[square,sort]{natbib}
\usepackage{float,graphicx,subfigure}
\usepackage{tikz}
\usepackage[skip=10pt]{caption}
\usepackage[breaklinks=true]{hyperref}
\usepackage{breakcites}
\usepackage{breakurl}

\theoremstyle{definition}
\newtheorem{theorem}{Theorem}[section]
\newtheorem{definition}[theorem]{Definition}
\newtheorem{lemma}[theorem]{Lemma}
\newtheorem{example}[theorem]{Example}

\newcommand{\Fig}[1]{Fig.~\ref{#1}}
\newcommand{\Tab}[1]{Tab.~\ref{#1}}
\newcommand{\Sec}[1]{Sect.~\ref{#1}}
\newcommand{\Eq}[1]{Eq.~(\ref{#1})}

\DeclareMathOperator{\Aut}{Aut}

\newcommand{\E}{\mathrm{e}}
\newcommand{\A}{\mathbf{A}}
\newcommand{\M}{\mathbf{M}}
\newcommand{\Z}{\mathbb{Z}}
\newcommand{\N}{\mathbb{N}}
\newcommand{\R}{\mathbb{R}}
\newcommand{\TT}[1]{\text{\tt #1}}

\renewcommand{\vec}[1]{\mathbf{#1}}

\bibpunct{(}{)}{,}{}{}{;}

\title{Invariants for neural automata}

\author[Uria-Albizuri et al.]{Jone Uria-Albizuri}
\address{ University of the Basque Country, Department of Mathematics, Leioa, Spain}
\email{jone.uria@ehu.eus}

\author[]{ Giovanni Sirio Carmantini}
\address{ foldAI, Munich, Germany}
\email{giovanni@carmantini.com}

\author[]{Peter beim Graben}
\address{Bernstein Center for Computational Neuroscience, Berlin, Germany }
\email{Peter.beimGraben@b-tu.de}

\author[]{Serafim Rodrigues}
\address{Basque Center for Applied Mathematics, Bilbao, Spain}
\email{srodrigues@bcamath.org}

\date{\today}

\begin{document}
\maketitle


\begin{abstract}
Computational modeling of neurodynamical systems often deploys neural networks and symbolic dynamics. One particular way for combining these approaches within a framework called \emph{vector symbolic architectures} leads to neural automata. An interesting research direction we have pursued under this framework has been to consider mapping symbolic dynamics (e.g. performed by Turing machines) onto neurodynamics, represented as neural automata. This representation theory, enables us to ask questions, such as, how does the brain implement Turing computations. Specifically, in this representation theory, neural automata result from the assignment of symbols and symbol strings to numbers, known as  G\"odel encoding. Under this assignment symbolic computation becomes represented by trajectories of state vectors in a real phase space, that allows for statistical correlation analyses with real-world measurements and experimental data. However, these assignments are usually completely arbitrary. Hence, it makes sense to address the problem question of, which aspects of the dynamics observed under such a representation is intrinsic to the dynamics and which are not. In this study, we develop a formally rigorous mathematical framework for the investigation of symmetries and invariants of neural automata under different encodings. As a central concept we define \emph{patterns of equality} for such systems. We consider different macroscopic observables, such as the mean activation level of the neural network, and ask for their invariance properties. Our main result shows that only step functions that are defined over those patterns of equality are invariant under symbolic recodings, while the mean activation, e.g., is not. Our work could be of substantial importance for related regression studies of real-world measurements with neurosymbolic processors for avoiding confounding results that are dependant on a particular encoding and not intrinsic to the dynamics.
\end{abstract}


\section{Introduction}
\label{sec:intro}
Computational cognitive neurodynamics deals to a large extent with statistical modeling and regression analyses between behavioral and neurophysiological observables on the one hand and neurocomputational models of cognitive processes on the other hand \citep{GazzanigaIvryMangun02, RabinovichFristonVarona12}. Examples for experimentally measurable observables are response times (RT), eye-movements (EM), event-related brain potentials (ERP) in the domain of electroencephalography (EEG), event-related magnetic fields (ERF) in the domain of magnetoencephalography (MEG), or the blood-oxygen-level-dependent signal (BOLD) in functional magnetic resonance imaging (fMRI).

Computational models for cognitive processes often involve drift-diffusion approaches \citep{Ratcliff78, RatcliffMcKoon07}, cognitive architectures such as ACT-R \citep{AndersonEA02}, automata theory \citep{HopcroftUllman79}, dynamical systems \citep{Gelder98, Kelso95, RabinovichVarona18}, and notably neural networks (e.g.~\citet{HertzKroghPalmer91, Arbib95}) that became increasingly popular after the induction of deep learning techniques in recent time \citep{CunBengioHinton15, Schmidhuber15}.

For carrying out statistical correlation analyses between experimental data and computational models one has to devise \emph{observation models}, relating the microscopic states within a computer simulation (e.g. the spiking of a simulated neuron) with the above-mentioned macroscopically observable measurements. In decision making, e.g.,~a suitable observation model is first passage time in a drift-diffusion model \citep{Ratcliff78, RatcliffMcKoon07}. In the domain of neuroelectrophysiology, local field potentials (LFP) and EEG can be described through macroscopic mean-fields, based either on neural compartment models \citep{MazzoniPanzeriEA08, GrabenRodrigues13, MartinezEA21}, or neural field theory \citep{JirsaEA02, GrabenRodrigues14}. For MRI and BOLD signals, particular hemodynamic observation models have been proposed \citep{FristonEA00, StephanEA04}.

In the fields of computational psycholinguistics and computational neurolinguistics \citep{ArbibCaplan79, Crocker96, GrabenDrenhaus12, Lewis03} a number of studies employed statistical regression analysis between measured and simulated data. To name only a few of them, \Citet{DavidsonMartin13} modeled speed-accuracy data from a translation-recall experiment among Spanish and Basque subjects through a drift-diffusion approach \citep{Ratcliff78, RatcliffMcKoon07}. \Citet{LewisVasishth06} correlated self-paced reading times for English sentences of different linguistic complexity with the predictions of an ACT-R model \citep{AndersonEA02}. \Citet{Huyck09} devised a Hebbian cell assembly network of spiking point neurons for a related task. Using an automaton model for formal language \citep{HopcroftUllman79}, \citet{Stabler11b} argued how reading times could be related to the automaton's working memory load. Similarly, \citet{BostonEA08} compared eye-movement data with the predictions of an automaton model for probabilistic dependency grammars \citep{Nivre08}.

Correlating human language processing with event-related brain dynamics became an important subject of computational neurolinguistics in recent years. Beginning with the seminal studies of \citet{GrabenLiebscherSaddy00, GrabenJurishEA04}, similar work has been conducted by numerous research groups (for an overview cf.~ \citet{HaleCampanelliEA22}). Also to name only a few of them, \citet{HaleLutzEA15} correlated different formal language models with the BOLD response of participants listening to speech. Similarly, \citet{FrankEA15} used different ERP components in the EEG, such as the N400 (a deflection of negative polarity appearing about 400 ms after stimulus onset as a marker of lexical-semantic access) for such statistical modeling. \Citet{GrabenDrenhaus12} correlated the temporally integrated ERP during the understanding of negative polarity items \citep{Krifka95} with the harmony observable of a recurrent neural network \citep{Smolensky06}, thereby implementing a formal language processor as a \emph{vector symbolic architecture} \citep{Gayler06, SchlegelNeubertProtzel21}. Another neural network model of the N400 ERP-component is due to \citet{RabovskyMcRae14}, and to \citet{RabovskyHansenMcClelland18} who related this marker with neural prediction error and semantic updating as observation models. Similar ideas have been suggested by \citet{BrouwerCrockerEA17, BrouwerCrocker17}, and \citet{BrouwerDeloguEA21} who considered a deep neural network of layered simple recurrent networks \citep{CleeremansServanMcClelland89, Elman90}, where the basal layer implements lexical retrieval, thus accounting for the N400 ERP-component, while the upper layer serves for contextual integration. Processing failures at this level are indicated by another ERP-component, the P600 (a positively charged deflection occurring around 600 ms after stimulus onset). Their neurocomputational model thereby implemented a previously suggested retrieval-integration account \citep{BrouwerFitzHoeks12, BrouwerHoeks13}.

In the studies of \citet{GrabenLiebscherSaddy00, GrabenJurishEA04, GrabenGerthVasishth08}, a dynamical systems approach was deployed --- later dubbed \emph{cognitive dynamical modeling} by \citet{GrabenPotthast09a}. This denotes a three-tier approach starting firstly with symbolic data structures and algorithms as models for cognitive representations and processes. These symbolic descriptions are secondly mapped onto a vectorial representation within the framework of vector symbolic architectures \citep{Gayler06, SchlegelNeubertProtzel21} through filler-role bindings and subsequent tensor product representations \citep{Smolensky90, Smolensky06, Mizraji89, Mizraji20}. In a third step, these linear structures are used as training data for neural network learning. More specifically, symbol strings and formal language processors can be mapped through G\"odel encodings to \emph{dynamical automata} \citep{GrabenLiebscherSaddy00, GrabenJurishEA04, GrabenGerthVasishth08, GrabenPotthast14}. Quite recently, \citet{CarmantiniEA17} have demonstrated how to realize those devices parsimoniously as modular recurrent neural networks, called \emph{neural automata} (NA), henceforth.\footnote{
    Note that neural automata are parsimonious implementations of universal computers, especially of Turing machines. These are not to be confused with neural Turing machines appearing in the framework of deep learning approaches \citep{GravesWayneDanihelka14}.
}
\Citet{CarmantiniEA17} also showed how neural automata can be used for neurolinguistic correlation studies. They implemented a diagnosis-repair parser \citep{Lewis98, LewisVasishth06} for the processing of initially ambiguous subject relative and object relative sentences \citep{FrischGrabenSchlesewsky04, LewisVasishth06} through an interactive automata network. As an appropriate observation model they exploited the mean activation of the resulting neural network \citep{Amari74} as \emph{synthetic ERP} \citep{GrabenGerthVasishth08, BarresSimonsArbib13} and obtained a model for the P600 component in their attempt.

For all these neurocomputational models symbolic content must be encoded as neural activation patterns. In vector symbolic architectures, this procedure involves a mapping of symbols onto filler vectors and of their possible binding sites in a data structure onto role vectors \citep{GrabenPotthast09a}. Obviously, such an encoding is completely arbitrary and could be replaced at least by any possible permutation of a chosen code. Therefore, the question arises to what extent neural observation models remain \emph{invariant} under permutations of an arbitrarily chosen code. Even more crucially, one has to face the problem whether a statistical correlation analysis depends on only one particularly chosen encoding, or not. Only if statistical models are also invariant under recoding, they could be regarded as reliable methods of scientific investigation.

It is the aim of the present study to provide a rigorous mathematical treatment of invariant observation models for the particular case of dynamical and neural automata and their underlying shift spaces. The article is structured as follows. In \Sec{sec:dsinv} we introduce the general concepts and basic definitions about invariants in dynamical systems, focusing later in \Sec{sec:neudy} on the special case of neurodynamical ones. In \Sec{sec:sydy} we focus our attention on symbolic dynamics. After introducing the basic notation we introduce the tools and facts that are needed in \Sec{sec:rtree} about rooted trees and about G\"odel encodings in \Sec{sec:godel}. In \Sec{sec:cylset} we relate these concepts to cylinder sets in order to finally describe the invariant partitions for different G\"odelizations of strings in \Sec{sec:invariants}. Then, in \Sec{sec:na} we describe the architecture for  neural automata and how to pass from single strings to dotted sequences. Finally, in \Sec{sec:invobs} we describe a symmetry group defined by G\"odel recoding of alphabets for neural automata, and we define a macroscopic observable that is invariant under this symmetry, based on the invariants described in \Sec{sec:invariants} before. In the end, in \Sec{sec:nlapp}, we apply our results to a concrete example with a neural automaton constructed to emulate parser for a context-free grammar. We demonstrate that the given macroscopic observable is invariant under G\"odel recodings, whereas Amari's mean network activity is not. Section \ref{sec:disc} provides a concluding discussion. All the mathematical proofs about the facts claimed throughout the paper are collected in an appendix.


\section{Invariants in dynamical systems}
\label{sec:dsinv}
We consider a classical time-discrete and deterministic dynamical system in its most generic form as an ordered pair $\Sigma = (X, \Phi)$ where $X \subset \R^n$ is a compact Hausdorff space as its phase space of dimension $n \in \N$ and $\Phi: X \to X$ is an invertible (generally nonlinear) map \citep{AtmanspacherGraben07}. The flow of the system is generated by the time iterates $\Phi^t$, $t \in \Z$, i.e., $t \mapsto \Phi^t$ is a one-parameter group for the dynamics with time $t \in \Z$, obeying $\Phi^t \circ \Phi^s = \Phi^{t + s}$ for $t, s \in \Z$. The elements of the phase space $\vec{x} \in X$ refer to the microscopic description of the system $\Sigma$ and are therefore called \emph{microstates}. After preparation of an \emph{initial condition} $\vec{x}_0 \in X$ the system evolves deterministically along a \emph{trajectory} $T = \{ \vec{x}(t) = \Phi^t(\vec{x}_0) | \, t \in \Z\}$.

A bounded function $f : X \to \R$ is called an \emph{observable} with $f(\vec{x}) \in \R$ as measurement result in microstate $\vec{x}$. The function space $B(X) = \{ f : X \to \R |  \, \|f\| < \infty \}$, endowed with point wise function addition $(f + g)(x) = f(x) + g(x)$, function multiplication $(f  g)(x) = f(x) g(x)$, and scalar multiplication $(\lambda f)(x) = \lambda f(x)$ (for all $f, g \in B(X)$, $\lambda \in \R$) is called the observable algebra of the system $\Sigma$ with norm $\|\cdot\| : B(X) \to \R_0^+$. Restricting the function space $B(X)$ to the bounded continuous functions $C_0(X)$, yields the algebra of \emph{microscopic observables} which describe ideal measurements for uniquely distinguishing among different microstates within certain regions of phase space.

By contrast, complex real-world dynamical systems only allow the measurement of macroscopic properties. The corresponding \emph{macroscopic observables} belong to the larger algebra of bounded functions\footnote{
    In fact one needs the algebra of essentially bounded functions with respect to a given probability measure here. For a proper treatment of these concepts, \emph{algebraic quantum theory} is required \citep{Sewell02}.
}
$B(X)$ and are usually defined as large-scale limits of so-called mean-fields \citep{Hepp72, Sewell02}. Examples for macroscopic mean-field observables in computational neuroscience are discussed below.

The algebra of macroscopic observables $B(X)$ contains step functions and particularly the indicator functions $\chi_A$ for proper subsets $A \subset X$ which are not continuous over whole $X$. Because $\chi_A(\vec{x}) = \chi_A(\vec{y})$ for all $\vec{x}, \vec{y} \in A$, the microstates  $\vec{x}$ and $\vec{y}$ are not distinguishable by means of the macroscopic measurement of $\chi_A$. Thus, \citet{Jauch64} and \citet{Emch64} called them \emph{macroscopically equivalent}.\footnote{
    Cf. the related concept of epistemic equivalence used by \citet{GrabenAtmanspacher06a, GrabenAtmanspacher09}.
}
The class of macroscopically equivalent microstates forms a \emph{macrostate} in the given mathematical framework \citep{Jauch64, Emch64, Sewell02}. Hence, a macroscopic observable induces a partition of the phase space of a dynamical system $\Sigma$ into macrostates.

The algebras of microscopic observables, $C_0(X)$, and of macroscopic observables, $B(X)$, respectively, are linear spaces with their additional algebraic products. As vector spaces, they allow the construction of linear homomorphism $\varphi: B(X) \to B(X)$ which are vector spaces either. An important subspace of the space of linear homomorphism is provided by the space of linear automorphisms, $\Aut(B(X))$, which contains the invertible linear homomorphisms. The space $\Aut(B(X))$ is additionally a group with respect to function composition, $(\varphi \circ \eta)(f)$, called the \emph{automorphism group} of the algebra $B(X)$.

Next, let $G$ be a group possessing a faithful representation $\alpha$ in the automorphism group $\Aut(B(X))$ of the dynamical system $\Sigma$. Then, for $a \in G$, $\alpha_a \in \Aut(B(X))$ maps an observable $f \in B(X)$ onto its transformed $\alpha_a(f) \in B(X)$, such that for two $a, b \in G$ it holds $\alpha_{a * b}(f) = (\alpha_a \circ \alpha_b)(f)$ where `$*$' denotes the group product in $G$. The group $G$ is called a \emph{symmetry} of the dynamical system $\Sigma$ \citep{Sewell02}. Moreover, if the representation of $G$ commutes with the dynamics of $\Sigma$,
\begin{equation}\label{eq:dynsym}
    (\alpha_a(f \circ \Phi^t))(\vec{x}) = f(\Phi^t(\alpha_a^*(\vec{x})))
\end{equation}
for all $\vec{x} \in X$, the group $G$ is called \emph{dynamical symmetry} \citep{Sewell02}. In \Eq{eq:dynsym}, the map $\alpha_a^*: X \to X$ is the lifting result from the observables to phase space through
\begin{equation}\label{eq:lift}
    f \circ \alpha_a^* = \alpha_a(f) \:.
\end{equation}

As an example consider the macroscopic observable $\chi_A$, i.e. the indicator function for a proper subset $A \subset X$ again. Choosing $\alpha_a^*$ in such as way that $\alpha_a^*(\vec{x}) \in A$ for all $\vec{x} \in A$, leaves $\chi_A$ invariant: $\chi_A(\alpha_a^*(\vec{x})) = \chi_A(\vec{x})$.

More generally, we say that an observable $f \in B(X)$ is \emph{invariant} under the symmetry $G$ if
\begin{equation}\label{eq:invari}
    f(\alpha_a^*(\vec{x})) = f(\vec{x})
\end{equation}
for all $a \in G$. It is the aim of the present study to investigate such invariants for particular neurodynamical systems, namely dynamical and neural automata \citep{GrabenLiebscherSaddy00, GrabenJurishEA04, GrabenGerthVasishth08, CarmantiniEA17}.


\subsection{Neurodynamics}
\label{sec:neudy}
Neurodynamical systems are essentially recurrent neural networks consisting of a large number, $n \in \N$, of model neurons (or units) that are connected in a complex graph \citep{HertzKroghPalmer91, Arbib95, CunBengioHinton15, Schmidhuber15}. Under a suitable normalization, the activity of a unit, e.g.~its spike rate can be represented by a real number in the unit interval $[0, 1] \subset \R$. Then, the microstate of the entire network becomes a vector in the $n$-dimensional hypercube, $\vec{x} \in X = [0, 1]^n \subset \R^n$. The microscopic observables are projectors on the individual coordinate axes,
\[
    f_i(\vec{x}) = x_i
\]
for $1 \le i \le n$. For discrete time, the network dynamics is generally given as a nonlinear difference equation
\begin{equation}\label{eq:netevolve}
    \vec{x}(t + 1) = \Phi_\vec{W}(\vec{x}(t)) \:.
\end{equation}
Here $\vec{x}(t) \in X$ is the activation vector (the microstate) of the network at time $t$ and $\Phi_\vec{W}$ is a nonlinear map, parameterized by the synaptic weight matrix $\vec{W} \in \R^{n^2}$. Often, the map $\Phi_\vec{W}$  is assumed to be of the form
\begin{equation}\label{eq:netevolve2}
    \Phi_\vec{W}(\vec{x}) = \vec{F}(\vec{W} \cdot \vec{x}) \:,
\end{equation}
with a nonlinear squashing function $\vec{F} = (F_i)_{1 \le i \le n}: X \to X$ as the \emph{activation function} of the network. For $F_i = \Theta$ (where $\Theta$ denotes the Heaviside jump function), equations (\ref{eq:netevolve}, \ref{eq:netevolve2}) describe a network of McCulloch-Pitts neurons \citep{McCullochPitts43}. Another popular choice for the activation function is the logistic function
\[
    F_i(x) = \frac{1}{1 + \E^{-x_i}} \:,
\]
describing firing rate models (cf., e.g., \citet{Graben08a}). Replacing \Eq{eq:netevolve2} by the map
\begin{equation}\label{eq:linetzdis}
    \Phi_\vec{W}(\vec{x}) = (1 - \Delta t) \vec{x} + \Delta t \, \vec{F}(\vec{W} \cdot \vec{x})
\end{equation}
yields a time-discrete leaky integrator network \citep{WilsonCowan72, GrabenBarrettAtmanspacher09, GrabenRodrigues13}. For numerical simulations using the Euler method, $\Delta t < 1$ is chosen for the time step.

For correlation analyses of neural network simulations with experimental data from neurophysiological experiments one needs a mapping from the high-dimensional neural activation space $X \subset \R^n$ into a much lower-dimensional \emph{observation space} that is spanned by $p \in \mathbb{N}$ macroscopic observables $f_k: X \to \R$ ($1 \le k \le p$). A standard method for such a projection is principal component analysis (PCA) \citep{Elman91}. If PCA is restricted to the first principal axis, the resulting scalar variable could be conceived as a measure of the overall activity in the neural network. In the realm of computational neurolinguistics PCA projections were exploited by \citet{GrabenGerthVasishth08}.

Another important scalar observable, e.g. used by \citet{GrabenDrenhaus12} as a neuronal observation model, is Smolensky's harmony \citep{Smolensky86}
\begin{equation}\label{eq:harmony}
    H(t) = \vec{x}(t)^+ \cdot \vec{W} \cdot \vec{x}(t)
\end{equation}
with $\vec{x}^+$ as transposed activation state vector, and the synaptic weight matrix $\vec{W}$, above.

\Citet{BrouwerCrockerEA17} suggested the ``dissimilarity'' between the actual microstate and its dynamical precursor, i.e.
\begin{equation}
    D(t) = 1 - \frac{\vec{x}(t) \cdot \vec{x}(t - 1)}{\|\vec{x}(t)\| \|\vec{x}(t - 1)\|}
\end{equation}
as a suitable neuronal observation model.

In this study, however, we use Amari's mean network activity \citep{Amari74}
\begin{equation}\label{eq:amari}
    A(t) = \frac{1}{n} \sum_{i} x_i(t)
\end{equation}
as time-dependent ``synthetic ERP'' \citep{BarresSimonsArbib13, CarmantiniEA17} of a neural network.


\subsection{Symbolic dynamics}
\label{sec:sydy}
A symbolic dynamics arises from a time-discrete but space continuous dynamical system $\Sigma$ through a partition of its phase space $X$ into a finite family of $m$ disjoint subsets totally covering the space $X$ \citep{LindMarcus95}. Hence
\[
    \mathcal{P} = \{ A_k \subset X | A_k \cap A_j = \emptyset \text{ for } k \ne j \:,\quad \bigcup_{k=1}^m A_k = X \} \:.
\]
Such a partition could be induced by a macroscopic observable with finite range. By assigning the index $k$ of a partition set $A_k$ as a distinguished \emph{symbol} $s_t$ to a state $\vec{x}(t)$ when $\vec{x}(t) \in A_k$, a trajectory of the system is mapped onto a two-sided infinite symbolic sequence. Correspondingly, the flow map of the dynamics $\Phi$ becomes represented by the left shift $\sigma$ through $\sigma(s_t) = s_{t + 1}$.

Following \citet{GrabenJurishEA04, GrabenGerthVasishth08}, and \citet{CarmantiniEA17}, a symbol is meant to be a distinguished element from a finite set $\A$, which we call an \emph{alphabet}. A sequence of symbols $w \in \A^l$ is called a word of length $l$, denoted $l = |w|$. The set of words of all possible lengths $w$ of finite length $|w| \ge 0$, also called the \emph{vocabulary} over $\A$, is denoted $\A^*$ (for $|w| = 0$, $w = \epsilon$ denotes the ``empty word'').


\subsubsection{Rooted trees}
\label{sec:rtree}
One can visualize the set of all words over the alphabet $\A$ as a regular rooted tree, $T$, where each vertex is labeled by and corresponds to each word formed by using this alphabet. Let us assume that $\A$ has $m$ letters for some $m\in\N$. That is $\A=\{a_1,\dots,a_m\}$. Then, the tree $T$ is inductively constructed as follows:

\begin{itemize}
	\item[(i)] The root of the tree is a vertex labeled by the empty word $\epsilon$.
	\item[(ii)] Assume we have constructed the vertices of step $n$, then we construct the vertices of step $n+1$ as follows. Suppose that we have $k$ vertices at step $n$ that are labeled by the words $w_1,\dots,w_k$. Then
	\begin{itemize}
		\item[$\bullet$] For each $i=1,\dots,k$ and each $a_j\in \A$ we add a new vertex decorated by $w_ia_j$.
		\item[$\bullet$] For each $i=1,\dots,k$ and $j=1,\dots,m$ we add and edge from $w_i$ to $w_ia_j$.
	\end{itemize}
\end{itemize}
	
\begin{figure}[H]
	\centering
\includegraphics{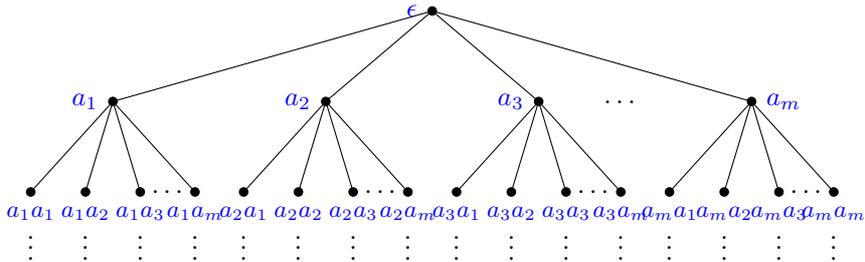}
\caption{The vocabulary $\A^*$ as a rooted tree.}\label{fig:rooted_tree_A}	
\end{figure}

This construction generates a regular rooted tree. Following the aforementioned construction, typically in the first step the root is placed at the top vertex. Subsequently the root is joined by edges, where each edge is associated to every word of length 1, that is, to every symbol of $\A$. Then iteratively, each of these edges labeled by a letter of $\A$ is joined to any word of length two starting by that letter, and so on. Assuming that $\A=\{a_1,\dots,a_m\}$, this construction yields an infinite tree as in \Fig{fig:rooted_tree_A}.

Each vertex of the tree corresponds to a word over the alphabet $\A$. That is, the set of vertices of the tree is $\A^*$. On the other hand, each infinite ray starting from the root, corresponds to an infinite sequence of symbols over $\A$, and it corresponds to the boundary of the tree. We denote this boundary by $\partial T$ and as mentioned, viewed as a set is equal to $\A^{\N}$.

The construction of the tree is unique up to the particular ordering of the symbols in $\A$ we chose. Thus, in principle, if $\gamma:\A\to\{0,\dots,m-1\}$ is a particular ordering (i.e. a bijection) of the alphabet where an element $a$ is denoted as $a_i$ if $\gamma(a)=i-1$, then the tree should be denoted by $T_\gamma$ as it depends on that particular ordering of the alphabet.

\begin{figure}[H]
    \centering
	\includegraphics{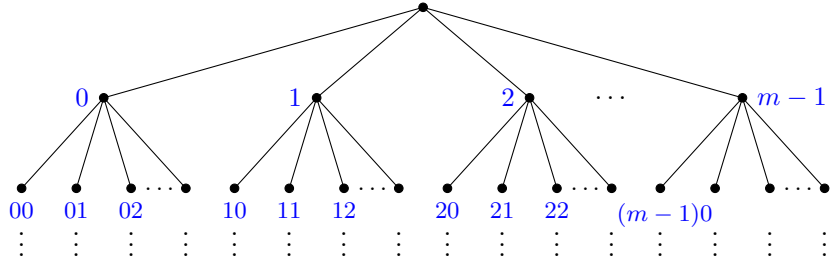}
	\caption{The regular rooted tree $T$ over the alphabet $\{0,1,\dots, m-1\}$.}\label{fig:rooted_tree_T}
\end{figure}

Let us denote by $T$ the regular rooted tree over the alphabet $\{0,1,\dots m-1\}$ with the natural order induced by $\N$ (see \Fig{fig:rooted_tree_T}).

Henceforth will denote by $\M$ the alphabet $\{0,\dots,m-1\}$ and as before, by $T$ the tree corresponding to the alphabet $\M$ under the usual ordering on $\N$.

When we say that the construction is unique up to reordering of symbols, we mean that both trees are isomorphic as graphs, where an isomorphism of graphs is a bijection between vertices preserving incidence. Indeed, for any bijection $\gamma:\A \to \M$, the tree $T_{\gamma}$ is ismorphic to $T$ as a graph.

\begin{lemma}\label{lem:iso}
	Let $\gamma:\A\to \M$ be an ordering of the alphabet $\A$. Then $T_{\gamma}$ and $T$ are isomorphic.
\end{lemma}

Since being isomorphic is transtive, this lemma shows that for any two alphabets $\A_1$ and $\A_2$ of the same cardinality and any two orderings of those alphabets $\gamma_1$ and $\gamma_2$, their corresponding trees $T_{\gamma_1}$ and $T_{\gamma_2}$ will be isomorphic as graphs.


\subsubsection{G\"odel encodings}
\label{sec:godel}
Having $\A^\N$, the space of one-sided infinite sequences over an alphabet $\A$ containing $\vert\A\vert =m$ symbols and $s=a_1 a_2 \ldots$ a sequence in this space, with $a_k$ being the $k$-th symbol in $s$ and an ordering $\gamma:\A\to\{0,\dots,m-1\}$, then a G\"{o}delization is a mapping from $\A^\N$ to $[0,1]\subset \R$ defined as follows:
\begin{equation}
	\psi_\gamma(s) := \sum\limits_{k=1}^{\infty} \gamma(a_k)m^{-k}.
	\label{eq:godel_encoding}
\end{equation}

By the Lemma \ref{lem:iso} we know that for each G\"odelization of $\A$ induced by $\gamma$, there is an isomorphism of graphs between $T_{\gamma}$ and $T$. Since the choice for the ordering of the alphabet (in other words, the choice of $\gamma$) is arbitrary and leads to different G\"odel encodings, we are interested in finding invariants for different such encodings.

One can define a metric on the boundary of the tree in the following way:
given any two infinite rays of the tree $p=a_1a_2a_3\dots$ and $q=b_1b_2b_3\dots$ we define
$$d(p,q)=\begin{cases}
	0 &\text{ if } p=q\\
	m^{-n}, &\text{ if } a_i=b_i \text{ for } i=1,\dots,n \text{ and } a_{n+1}\neq b_{n+1}\\
	1 &\text{ if } a_1\neq b_1.
\end{cases}$$

This defines an ultrametric on the boundary, that is, a metric that satisfies a stronger version of the triangular inequality, namely: $$d(p,q)\leq\max\{d(p,r),d(r,q)\}.$$ When we encode the infinite strings under the G\"odel encoding, we are sending rays that are close to each other under this ultrametric to points that are close in the $[0,1]$ interval under the usual metric.

\begin{lemma}\label{lem:distance}
Let $p=a_1a_2a_3\dots$ and $q=b_1b_2b_3\dots$ be two infinite strings over $\A$. Then for any G\"odel encoding $\psi$ we have that
	$$d(p,q)\leq \frac{1}{m^n}\iff \exists k\in\{0,\dots, m^n-1\}, \psi(p),\psi(q)\in \left[\frac{k}{m^n},\frac{k+1}{m^n}\right).$$
\end{lemma}

Recall  that the lemma does not mean that points that are close (with respect to the usual metric) on the $[0,1]$ interval come from rays that were close on the tree. For example, if the alphabet has $3$ letters, the points $1/3-\epsilon$ and $1/3$ are as close as we want for any $\epsilon>0$ but are at distance $0$ from each other on the tree. In fact, it gives a partition of the interval for each $n\in\N$ in a way that, if two points representing an infinite string are in the same interval according to the partition of the corresponding $n$, then they come from two rays that share at least a common prefix of length $n$.


\subsubsection{Cylinder sets}
\label{sec:cylset}
In symbolic dynamics, a cylinder set \citep{McMillan53} is a subset of the space $\A^\N$ of infinite sequences from an alphabet $\A$ that agree in a particular building block of length $l \in \N$. Thus, let $w = \A^*$ be a finite word $a_1a_2 \dots a_l$ of length $l$, we define the cylinder set
\begin{equation}\label{eq:cylinder}
	[w] = [a_1a_2 \dots a_l] =
	\{ s \in \A^{\N} \,| \, s_{k} =
	a_k , \quad k = 1, \dots, l \} \:.
\end{equation}

\begin{figure}[H]
\centering
	\includegraphics{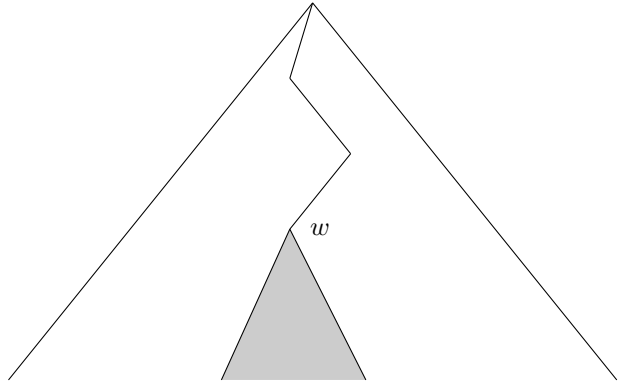}
	\caption{Cylinder set corresponding to $w$ seen on the tree.}\label{fig:rooted_tree_C}
\end{figure}

We can also see the cylinder sets on the tree depicted in \Fig{fig:rooted_tree_C}. In fact, for each level on the tree (where level refers to vertices corresponding to words of certain fixed length) we get a partition of the interval $[0,1]$. The vertices hanging from each vertex on that level land on their corresponding interval of the partition. Thus, from a rooted tree view point, a cylinder set corresponds to a whole tree hanging from that vertex. Concretely, the cylinder set $[w]$ for the word $w\in \A^*$ is the subtree hanging from the vertex decorated by $w$.

Two different G\"odel codes $\psi, \varphi$ can only differ with respect to their assignments  $\gamma_1, \gamma_2 : \A \to \{0,\dots,m-1\}$. Thus, we call a permutation $\pi \in S_m$ (with $S_m$ as the symmetric group) a \emph{G\"odel recoding}, if
\[
\pi \circ \gamma_1 = \gamma_2 \:.
\]


\subsubsection{Invariants}
\label{sec:invariants}
The ultimate goal of our study is to find invariants under G\"odel recodings. Observe that under the notation of Lemma \ref{lem:iso}, $g_{\gamma_1}:T_{\gamma_1}\to T$ and $g_{\gamma_2}:T_{\gamma_2}\to T$ are two graph isomorphisms. In fact, they induce a graph automorphism of $T$, $g_{\pi}=g_{\gamma_2}\circ g_{\gamma_1}^{-1}:T\to T$. And this automorphism sends the vertices encoded by $\gamma_1$ to the ones encoded by $\gamma_2$.

As Lemma \ref{lem:distance} shows, a G\"odel recoding preserves the size of cylinder sets after permuting vertices. However, the way of ordering the alphabet and how this permutes the rays of the tree is even more restrictive than just preserving the size of the cylinder sets. In fact, under the action of a reordering each vertex can only be mapped to certain vertices and it is forbidden to be sent to others. This is captured by the following most central definition.

\begin{definition}\label{def:patteq}
Let $w=a_{i_1}a_{i_2}\dots a_{i_l}\in \M^l$ be a string of length $l$ after an ordering $\gamma$. We define a partition of the set of integers $\{1,2,\dots,l\}$,
\begin{equation}\label{eq:inpart}
  \mathcal{P}_w = \{ \{j_1,\dots,j_k\} \subset \mathbb{N} | a_{i_{j_1}}=\dots=a_{i_{j_k}} \} \:.
\end{equation}
For any word $w\in \M^*$ we call $\mathcal{P}_w$ \emph{the pattern of equality of $w$}.
\end{definition}

Equipped with aforementioned formalisms we are now in a position to formulate the first main finding of our study as follows.
\begin{theorem}\label{thm:partition}
For any other vertex $u\in T$ there exists a G\"odel recoding $\pi$ such that $g_{\pi}(w)=u$ if and only if
\begin{eqnarray}
\label{eq:partition_cond1}
  u &\in& \M^l \\
  \label{eq:partition_cond2}
  P_w &=& P_u \:.
\end{eqnarray}
\end{theorem}

Theorem \ref{thm:partition} states that each vertex can be mapped to any vertex having the same pattern of equality and nowhere else.

\begin{example}
If $\A=\{a,b,c\}$ and we consider $w=aaabcabc\in \A^8$. Then we have $P_w=\{\{1,2,3,6\},\{4,7\},\{5,8\}\}$, which gives us all the possible words where $w$ can be mapped to. That would be the list of all the possibilities:
	$$	\begin{matrix}
		bbbacbac&cccbacba&aaacbacb\\
		bbbcabca&cccabcab&aaabcabc
	\end{matrix}$$
So we have only $6$ possible vertices out of $3^8=6561$. And of course, this proportion reduces as we go deeper on the tree.
\end{example}

\begin{figure}[H]
\centering
	\includegraphics{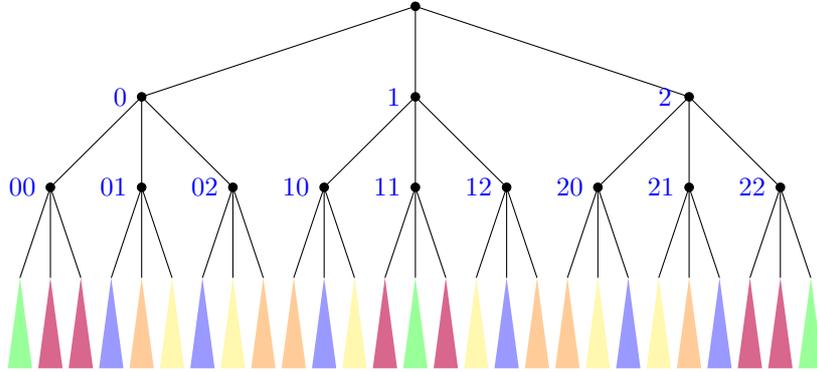}
	\caption{Invariant partition of the cylinder sets according to their patterns of equality.}\label{fig:invpart}
\end{figure}

In terms of G\"odelization  into the $[0,1]$ interval, we ilustrate the implications by an example. Let us assume that $m=3$ and $l=3$, for example. Then, in \Fig{fig:invpart} the cylinder sets of certain color can only be mapped through a recoding to a cylinder set of the same color and nowhere else.

\begin{figure}[H]
\centering
	\includegraphics{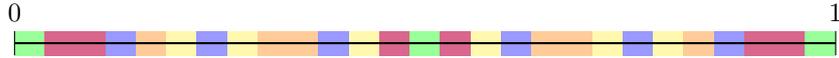}
	\caption{Invariant partition of the interval $[0,1]$ after G\"odelization.}\label{fig:partition_interval}
\end{figure}

Figure \ref{fig:partition_interval} shows the corresponding partition of the interval $[0,1]$ where the intervals in each color may be mapped to another of the same color by a different assignment map and nowhere else.

\subsection{Neural automata}
\label{sec:na}
Following \citet{GrabenJurishEA04, GrabenGerthVasishth08}, and \citet{CarmantiniEA17}, a \emph{dotted sequence} $s \in \A^\Z$ on an alphabet $\A$ is a two-sided infinite sequence of symbols ``$s = \ldots \; a_{-2} \; a_{-1} \; . \; a_{0} \; a_{1} \; a_{2} \; \ldots$'' where $a_i \in \A$, for all indices $i \in \Z$. Here, the dot ``.'' is simply used as a mnemonic sign, indicating that the index 0 is to its right.

\Citet{CarmantiniEA17} interpreted the dot as a meta-symbol which can be concatenated with two words $v_1, v_2 \in \A^*$ through $v = v_1 . v_2$. Let $\hat{\A}^*$ denote the set of these dotted words. Moreover, let $\Z^{-} = \{i\; |\; i < 0, \; i \in \Z \}$ and $\Z^{+} = \{ i\; | \; i \ge 0, \; i \in \Z \}$ the sets of negative and non-negative indices. We can then reintroduce the notion of a dotted sequence as follows. Let $s \in \A^\Z$ be a bi-infinite sequence of symbols such that $s = w_\alpha v w_\beta$ with $v \in \hat{\A}^*$ as a dotted word $v = v_1 . v_2$ and $w_\alpha v_1 \in \A^{\Z^{-}}$ and $v_2 w_\beta \in \A^{\Z^{+}}$. Through this definition, the indices of $s$ are inherited from the dotted word $v$ and are thus not explicitly prescribed.

A \emph{versatile shift} (VS) was defined by \citet{CarmantiniEA17} as a pair $M_{VS} = (\A^\Z, \Omega)$, with $\A^\Z$ being the space of dotted sequences, and $\Omega : \A^\Z \rightarrow \A^\Z$ defined by
\begin{equation}
	\Omega(s) = \sigma^{F(s)}(s \oplus G(s))
\end{equation}
with
\begin{equation}
	\begin{aligned}
		&F: \A^\Z \rightarrow \Z \\
		&\oplus: \A^\Z \times \A^\Z \rightarrow \A^\Z\\
		&G: \A^\Z \rightarrow \A^\Z,
	\end{aligned}
\end{equation}
where the operator ``$\oplus$'' substitutes the dotted word $v_1.v_2 \in \hat{\A}^*$ in $s$ with a new dotted word $\hat{v_1}.\hat{v_2} \in \hat{\A}^*$ specified by $G$, while $F(s) = F|_{\hat{\A}^*}(v_1.v_2)$ determines the number of shift steps as for Moore's generalized shifts \citep{CarmantiniEA17}.

A nonlinear dynamical automaton (NDA) is a triple $M_{NDA} = (Y, \mathcal{P}, \Phi)$, where $\mathcal{P}$ is a rectangular partition of the unit square $Y={[0, 1]}^2 \subset \R^2$, that is
\begin{equation}
	\mathcal{P} = \{D^{(i,j)} \subset Y |~ 1 \le i \le m,\; 1 \le j \le n,\; \space m,n \in \N\},
	\label{Eq:partition}
\end{equation}
so that each cell is defined as $D^{(i,j)} = I_i \times J_j$, with $I_i, J_j \subset [0,1]$ being real intervals for each bi-index $(i,j)$, with $D^{(i,j)} \cap D^{(k,l)} = \varnothing$ if $(i,j) \neq (k,l)$, and $\bigcup_{i,j} D^{(i,j)} = Y$. The couple $(Y, \Phi)$ is a time-discrete dynamical system with phase space $Y$ and the flow $\Phi: Y \rightarrow Y$ is a piecewise affine-linear map such that $\Phi_{|D^{(i,j)}}:=\Phi^{(i,j)}$, with $\Phi^{(i,j)}$ having the following form:
\begin{equation}
	\Phi^{(i,j)}(\vec{y}) = \left(\begin{array}{c} a^{(i,j)}_ 1 \\ a^{(i,j)}_2 \end{array}\right) +
	\left(\begin{array}{cc} \lambda^{(i,j)}_1 & 0 \\ 0 & \lambda^{(i,j)}_2 \end{array}\right)
	\left(\begin{array}{c} y_1 \\ y_2 \end{array}\right) \:,
	\label{Eq:NDA_dynamics}
\end{equation}
with state vector $\vec{y} = (y_1, y_2)$. \Citet{CarmantiniEA17} have shown that using G\"odelization any versatile shift can be mapped to a nonlinear dynamical automaton. Therefore, one can reproduce the activity of a versatile shift on the unit square $Y$. In order to do so, the partition (\ref{Eq:partition}) is given by the so called Domain of Dependance (DoD). The Domain of Dependance is a pair $(l,r)\in\N\times \N$ which defines the length of the strings on the left and right hand side of the dot in a dotted sequence that is relevant for the versatile shift to act on the phase space. The dynamics of the versatile shift is completely determined by how the string looks like in each iteration on the Domain of Dependance. Then, if the domain is $(l,r)$ and if the alphabet $\A$ has size $m$, the partition of the unit square is given by $m^r$ intervals on the $y_1$ axis and $m^l$ intervals on the $y_2$ axis, corresponding to cells where the NDA is defined according to the versatile shift. Finally, a \emph{neural automaton} (NA) is an implementation of an NDA by means of a modular recurrent neural network \citep{CarmantiniEA17}.

The neural automaton comprises a phase space $X = [0, 1]^n$ where the two-dimensional subspace $Y = [0, 1]^2$ of the underlying NDA is spanned by only two neurons that belong to the machine configuration layer (MCL). The remainder $X \setminus Y$ is spanned by the neurons of the branch selection layer (BSL) and the linear transformation layer (LTL), both mediating the piecewise affine mapping \eqref{Eq:NDA_dynamics}. Having an NDA defined from a versatile shift, each rectangle on the partition is given by the DoD, and the action of the NDA on each rectangle depends on the particular G\"odel encoding of the alphabet $\A$ that has been chosen. We are interested in invariant macroscopic observables of such automata under different G\"odel encodings of the alphabet.

Since we are now interested on dotted sequences over an alphabet $\A$, instead of having an invariant partition of the interval $[0,1]$ as in \Fig{fig:partition_interval}, we will have an invariant partition of the unit square $Y=[0,1]^2$. That is, we will have a partition in rectangles where the machine might be at certain step of the dynamics or not. Each color in that partition gives all the possible places where a particular dotted sequence of certain right and left lengths could be under a different G\"odel encoding.

\begin{figure}[H]
\begin{center}
\includegraphics{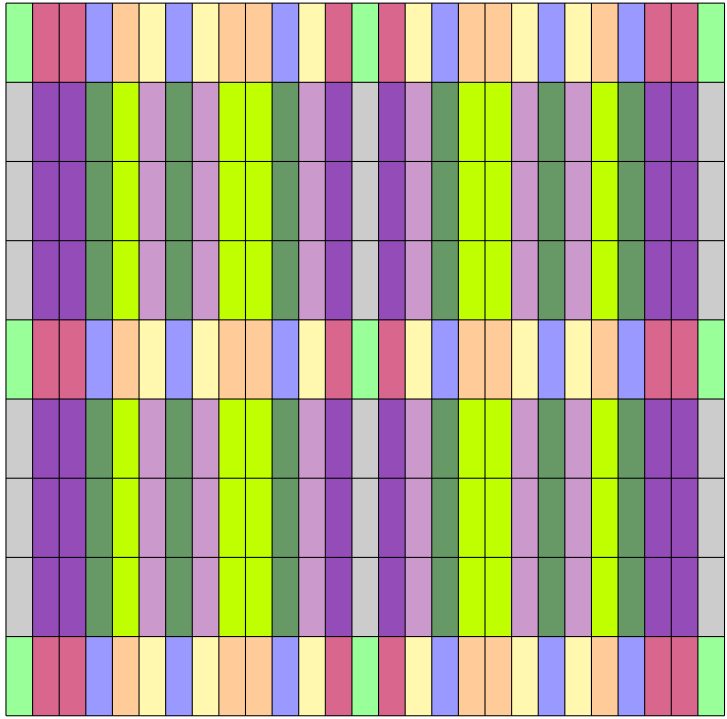}

\end{center}

\caption{Each small square corresponds to a square on the partition given by the dotted sequences of length $(2,3)$. The squares colored by the same color are those having the same pattern of equality, and thus, are those which can be mapped to each other under different G\"odel encodings of the alphabet.}\label{fig:invariant_partition_square}
\end{figure}

For example, assuming that our alphabet has $m = 3$ letters in both sides of the dotted sequence and that we are looking at words of length $l = 2$ on the left hand side of the dot, and length $r = 3$ on the right hand side of the dot, the partition would be like in Figure \ref{fig:invariant_partition_square}.

Let us assume that we are considering the invariant partition for dotted sequences of length $(l,r)$, meaning that the left hand side has length $l$ and the right hand side $r$. Then we know that the partition of the square $Y$ is given by $E^{(i,j)}=\left[\frac{i}{m^{l}},\frac{i+1}{m^{l}}\right)\times \left[\frac{j}{m^{r}},\frac{j+1}{m^{r}}\right)$. Each left corner of the rectangle corresponds to the position of the G\"odelization of a dotted sequence of size $(l,r)$. Each point $(y_1,y_2)=(\frac{i}{m^{l}},\frac{j}{m^l})$ has a unique expansion on base $m$ for its coordinates. That is, there are some $c_1,\dots,c_{l}$ with $0\leq c_i\leq m_1$ such that
\begin{equation}\label{Eq:point_to_sequence}
	y_1=\frac{i}{m^{l}}=\frac{c_1}{m}+\frac{c_2}{m^2}+\dots+\frac{c_{l}}{m^{l}}.
\end{equation}

These $\{c_1,\dots,c_{l}\}$ also define a partition of $\{1,\dots,l\}$ in the same way as given in definition \ref{def:patteq}. Therefore $\{d_1,\dots,d_k\}\in \mathcal{P}_x \iff c_{j_1}=\dots=c_{j_k}$. This procedure similarly applies to the $y_2$ coordinate. Hence, the corners defining an invariant piece of the partition will be those sharing the same partition of $\{1,\dots,l\}\times \{1,\dots,r\}$. In other words, we can obtain the corners related to a given $\vec{y}$ by expanding $y_1$ and $y_2$ on base $m$ and permuting the appearance of $0,\dots,m-1$ on the expansion.

For example, if $m=3$ and $(l,r)=(2,3)$, we have $3^2\cdot 3^3=3^5$ rectangles. Now let us take, for instance the rectangle $\left[\frac{6}{3^{2}},\frac{7}{3^{2}}\right)\times \left[\frac{10}{3^{3}},\frac{11}{3^{3}}\right)$ and let us find its invariant partition. First we decompose

$$y_1=\frac{6}{3^{2}}=\frac{2}{3}+\frac{0}{9}\,\,\,\text{and}\,\, y_2=\frac{10}{3^{3}}=\frac{1}{3}+\frac{0}{9}+\frac{1}{27}.$$

Hence a rectangle in the same invariant partition must be of the form $E^{(i,j)}=\left[\frac{i}{m^{l}},\frac{i+1}{m^{l}}\right)\times \left[\frac{j}{m^{r}},\frac{j+1}{m^{r}}\right)$ with $\frac{i}{3^2}=\frac{a}{3}+\frac{b}{9}$ and $\frac{j}{3^{3}}=\frac{c}{3}+\frac{b}{9}+\frac{c}{27}$ with $a,b,c\in\{0,1,2\}$ and different\footnote{Here we are assuming that both the alphabet $\A$ and the G\"odel Encoding is the same in both sides of the dot, otherwise we would have more freedom and obtain more intervals, but the procedure works anyway.}. This gives the following rectangles

\begin{align*}
	\left[\frac{1}{9},\frac{2}{9}\right)\times \left[\frac{23}{27},\frac{24}{27}\right) &&& \left[\frac{3}{9},\frac{4}{9}\right)\times \left[\frac{20}{27},\frac{21}{27}\right)\\
	\\&&&\\
	\left[\frac{2}{9},\frac{3}{9}\right)\times \left[\frac{16}{27},\frac{17}{27}\right) &&& \left[\frac{6}{9},\frac{7}{9}\right)\times \left[\frac{10}{27},\frac{11}{27}\right)\\
	\\&&&\\
	\left[\frac{7}{9},\frac{8}{9}\right)\times \left[\frac{3}{27},\frac{4}{27}\right) &&& \left[\frac{5}{9},\frac{6}{9}\right)\times \left[\frac{6}{27},\frac{7}{27}\right)\\
\end{align*}

In this way we can construct the partition of the unit square given by the patterns of equality.


\subsubsection{Invariant observables}
\label{sec:invobs}

Our aim now is to define an observable $f\in B(X)$, in the sense of Section \ref{sec:dsinv} for neural automata. That is $f:X\to \R$ should obey \Eq{eq:invari} where the map $\alpha^*_{\pi}$ corresponds to a symmetry induced by a G\"odel recoding of the alphabets. Here $\pi$ denotes the permutation of the alphabet needed to pass from one G\"odel encoding to the other, as explained later.

Notice that in the previous discussion we were assuming that we knew the length of the strings that were encoded. However, this is not the case in practice, and may cause problems, as the length of the strings vary at each iteration. For instance, if for the alphabet $\{a,b\}$ the symbol $a$ is mapped to $0$ under certain G\"odel enconding $\gamma$ and the symbol $b$ to $1$, then the number $x=1/2\in[0,1]$ would correspond to the word $w_r=ba\overset{r-1}{\dots}a$ once we assume that the string is of length $r$ for $r\in\N\cup\{0\}$. However, if we do not know the length of the encoded string, each $w_r$ will have a different G\"odel number under the G\"odel encoding that sends $b$ to $0$ and $a$ to $1$, namely $\sum_{k=2}^{r-1}1/2^{k}$. Thus, encoding symbols with the number $0$ makes some strings indistinguishible under G\"odel recoding, because having no symbols is interpreted as having the symbol encoded by $0$ as many times as we want. This issue can be easily avoided by adding one symbol $\sqcup$ to the alphabet, which will be interpreted as a blank symbol, and will always be forced to be encoded as $0$ by any G\"odel encoding.

Suppose that we have an NDA defined from a versatile shift under the condition that the blank symbol $\sqcup$ has been added to the alphabet $\A$ representing the blank symbol and that is mapped to $0$ under any G\"odel encoding\footnote{In order to make things simpler we will assume that we have the same alphabet on the stack and the input symbols. This can always be assumed considering the union of both alphabets if needed.}. We will assume that $\A$ has $m$ symbols after adding the blank symbol (that is, we had $m-1$ symbols before). Then for any pair $(r,l)\in\N\times\N$, we can divide the unit square $Y$ into the rectangle partition given by

\begin{equation}
	\mathcal{R} = \{E^{(i,j)} \subset Y |~ 1 \le i \le m^r,\; 1 \le j \le m^l\}.
	\label{Eq:partition_2}
\end{equation}

Next, we extend this partition of the phase space of the NDA, that equals the subspace of the machine configuration layer of the larger NA, to the entire phase space of the neural automaton. This is straightforwardly achieved by defining another partition
\begin{equation}
	\mathcal{Q} = \{ E^{(i,j)} \times [0,1]^{n-2} \subset X |~ 1 \le i \le m^r,\; 1 \le j \le m^l\}.
	\label{Eq:partition_3}
\end{equation}

Now, for each left corner $(x_1^{(i,j)}, x_2^{(i,j)})\in E^{(i, j)}$ we find their pattern of equality $\mathcal{P}_{ij}$, assuming that the permutation is taking place just on the symbols $\{2,\dots,m\}$ (as the first symbol has to be mapped to $0$ under any encoding).

Let us suppose that $\{\mathcal{P}_{ij_1},\dots, \mathcal{P}_{ij_s} \}$ are all the different appearing patterns of equality and we define the indicator functions $\chi_k: X \to \{0,1\}$ as
\begin{equation}\label{eq:indicator}
    \chi_k(\vec{x}) = \begin{cases}
                      1 & \text{if } \vec{x} \in E^{(i,j)} \times [0,1]^{n-2} \text{ and } \mathcal{P}_{ij}=\mathcal{P}_{ij_k} \\
                      0 & \text{otherwise}
                    \end{cases}
\end{equation}
for $\vec{x} \in X$. Then, we can choose $c_1,\dots,c_s\in\R$ to be $s$ different real numbers and define a macroscopic observable $f:X\to \R$ as a step function
\begin{equation}\label{eq:inv_f}
    f(\vec{x})=\sum_{k=1}^{s} c_k \chi_k(\vec{x}) \:.
\end{equation}
Clearly, we have $f \in B(X)$.

Our aim is to show that this observable is invariant under the symmetry group $S_{m-1}\times S_{m-1}$ of the dynamical system $(X,\Phi)$ given by the neural automaton in \Eq{Eq:NDA_dynamics}, where $S_{m-1}$ denotes the symmetric group on $m-1$ elements. First of all, we must show that $S_{m-1}\times S_{m-1}$ is a symmetry of the neural automaton.

Before doing this, we will define an auxiliary map. Let $\pi=(\pi_1,\pi_2)\in S_{m}\times S_{m}$ be any element of the product that fixes $1$ (on the set $\{1,2,\dots,m\}$ where $S_m$ acts). Notice that the elements of $S_m$ fixing the first element form a subgroup of $S_m$ that is isomorphic to $S_{m-1}$. Let now $\vec{x}$ be any point in $X$. Let us consider $\vec{y}=(y_1,y_2)$ the first two coordinates of $\vec{x}$ given by the activations of the machine configuration layer of the NA. Then, we can check in which of the intervals of the partition $\mathcal{R}$ is, say $(y_1,y_2)\in E^{(i,j)}=\left[\frac{i}{m^{l}},\frac{i+1}{m^{l}}\right)\times \left[\frac{j}{m^{r}},\frac{j+1}{m^{r}}\right)$. We can therefore compute the expansion on base $m$ of each corner and take the coefficients we get as words over the alphabet $\M=\{0,1,\dots,m-1\}$, say $c_1\dots c_l\in \M^l$ and $d_1,\dots,d_r\in \M^r$. Then, we compute $g_{\pi_1}(c_1\dots c_l)$ and $g_{\pi_2}(d_1\dots d_r)$ and we encode these words by the canonical G\"odel encoding (that is, the one given by the identity map on $\M$).  Thus, we obtain a new corner of some rectangle in our partition of the phase space, say  $E^{(i',j')}=\left[\frac{i'}{m^{l}},\frac{i'+1}{m^{l}}\right)\times \left[\frac{j'}{m^{r}},\frac{j'+1}{m^{r}}\right)$. We now define a map $\rho_{\pi}:Y\to Y$ by $\rho_{\pi}(y_1,y_2)=\left(y_1+\frac{i'-i}{m^{l}},y_2+\frac{j'-j}{m^{r}}\right)$. This map can obviously be extended to a map from $X$ to $X$ being the identity on the rest of the coordinates. Abusing notation we also refer to $\rho_{\pi}$ as to this map. Informally speaking, the map $\rho_{\pi}$ rigidly permutes the squares on the partition $\mathcal{R}$ according to the action of $g_{\pi_1}$ and $g_{\pi_2}$ on the words representing the corners.

Now, we can define $\alpha_{\pi}:B(X)\to B(X)$ as follows. For any $f\in B(X)$, we define
\begin{equation}\label{eq:nasymm}
    \alpha_{\pi}(f)(\vec{x})=f(\rho_{\pi}(\vec{x})) \:.
\end{equation}
It is not difficult to check that if $\pi,\gamma\in S_{m-1}\times S_{m-1}$ are two group elements, then $\alpha_{\gamma\pi}(f)=(\alpha_{\gamma}\circ\alpha_{\pi})(f)$ so that $S_{m-1}\times S_{m-1}$ is a symmetry of the system.

Thus, we obtain finally our main result.
\begin{theorem}\label{thm:f_invariant}
Let $f \in B(X)$ be a macroscopic observable on the space space of a neural automaton as defined in \eqref{eq:inv_f}. Then $f$ is invariant under the symmetric group $S_{m-1}\times S_{m-1}$ of G\"odel recodings of the automaton's symbolic alphabet.
\end{theorem}
It is worth mentioning that this procedure gives infinitely many different invariant observables. In fact, any choice of $(r,l)\in\N\times \N$ gives a thinner invariant partition, and respectively, a sharper observable.


\section{Neurolinguistic application}
\label{sec:nlapp}

As an instructive example we consider a toy model of syntactic language processing as often employed in computational psycholinguistics and computational neurolinguistics \citep{ArbibCaplan79, Crocker96, GrabenDrenhaus12, HaleCampanelliEA22, Lewis03}.

In order to process the sentence given by \citet{GrabenPotthast14} in example \ref{ex:sentence}, linguists often derive a \emph{context-free grammar} (CFG) from a phrase structure tree \citep{HopcroftUllman79}.
\begin{example}\label{ex:sentence}
    \TT{the dog chased the cat}
\end{example}
In our case, the CFG consists of \emph{rewriting rules}
\begin{align}
    \TT{S} &\to \TT{NP \ VP} \label{eq:PG:cfg1}\\
    \TT{VP} &\to \TT{V \ NP} \label{eq:PG:cfg2}\\
    \TT{NP} &\to \TT{the \ dog} \label{eq:PG:cfg3}\\
    \TT{V} &\to \TT{chased} \label{eq:PG:cfg4}\\
    \TT{NP} &\to \TT{the \ cat} \label{eq:PG:cfg5}
\end{align}
where the left-hand side always presents a nonterminal symbol to be expanded into a string of nonterminal and terminal symbols at the right-hand side. Omitting the lexical rules (\ref{eq:PG:cfg3} -- \ref{eq:PG:cfg5}), we regard the symbols $\TT{NP}, \TT{V}$, denoting `noun phrase' and `verb', respectively, as terminals and the symbols $\TT{S}$ (`sentence') and $\TT{VP}$ (`verbal phrase') as nonterminals.

Then, a versatile shift processing this grammar through a simple top down recognizer \citep{HopcroftUllman79} is defined by
\begin{equation}\label{eq:PG:gs}
\begin{array}{l @{\:\mapsto \:}l }
    \TT{S} . a & \TT{VP}\:\TT{NP} . a \\
    \TT{VP} . a & \TT{NP}\:\TT{V} . a \\
    a . a & \epsilon . \epsilon
\end{array}
\end{equation}
where the left-hand side of the tape is now called `stack' and the right-hand side `input'. In \eqref{eq:PG:gs} $a$ stands for an arbitrary input symbol. Note the reversed order for the stack left of the dot. The first two operations in \eqref{eq:PG:gs} are \emph{predictions} according to a rule of the CFG while the last one is an \emph{attachment} of subsequent input with already predicted material.

This machine then \emph{parses} the well formed sentence $\TT{NP}\,\,\TT{V}\,\,\TT{NP}$ as shown in Table 1 from \citet{GrabenPotthast14}. We reproduce this table here as \Tab{tab:PG:parse}.

\begin{table}[H]
\caption{\label{tab:PG:parse} Sequence of state transitions of the versatile shift processing the well-formed string from example \ref{ex:sentence}, i.e. \TT{NP V NP}. The operations are indicated as follows: ``predict (X)'' means prediction according to rule (X) of the context-free grammar; attach means cancelation of successfully predicted terminals both from stack and input; and ``accept'' means acceptance of the string as being well-formed.}
  \centering
\begin{tabular}{cr@{ . }ll}
  \hline
  time & \multicolumn{2}{c}{state} & operation \\
  \hline
  0 & \TT{S} & \TT{NP V NP} & predict \eqref{eq:PG:cfg1} \\
  1 & \TT{VP NP} & \TT{NP V NP} & attach \\
  2 & \TT{VP} & \TT{V NP} & predict \eqref{eq:PG:cfg2}  \\
  3 & \TT{NP V} & \TT{V NP} & attach  \\
  4 & \TT{NP} & \TT{NP} & attach  \\
  5 & $\epsilon$ & $\epsilon$  & accept \\
  \hline
\end{tabular}
\end{table}

\begin{figure}[H]
\includegraphics{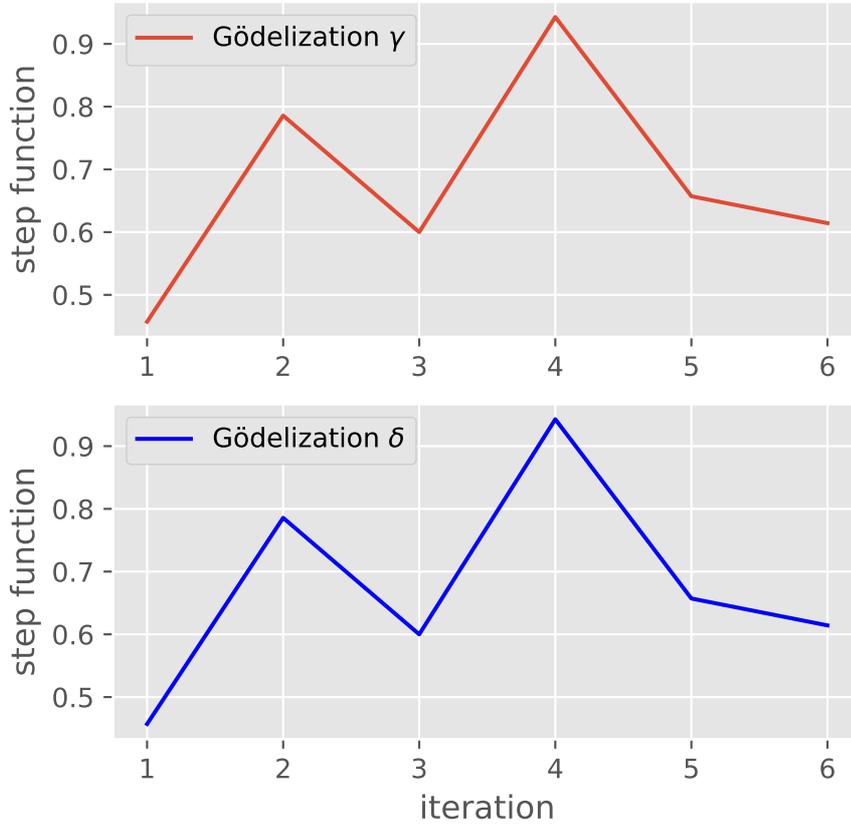}
\caption{The macroscopic observable $f$, given by the step function \eqref{eq:inv_f} is invariant under G\"odel recoding. The figure shows the result of `measuring' $f$ to a neural automaton encoded by $\gamma$ on top and to the same machine encoded by $\delta$ below.}\label{fig:step_function}
\end{figure}

Once we obtained the versatile shift, an NA simulating it can be generated. When we do so, we chose a particular G\"odel encoding of the symbols. Suppose we chose the following two G\"odelizations $\gamma=(\gamma_1,\gamma_2)$ and $\delta=(\delta_1,\delta_2)$ that are given by
\begin{align*}
		\gamma_1:\{\sqcup,\TT{NP},\TT{V}\}&\to \{0,1,2\}\\
		\sqcup&\mapsto 0\\
		\TT{NP}&\mapsto 1\\
		\TT{V}&\mapsto 2 \\
		\gamma_2:\{\sqcup,\TT{NP},\TT{V},\TT{VP},\TT{S}\}&\to \{0,1,2,3,4\}\\
		\sqcup&\mapsto 0\\
		\TT{NP}&\mapsto 1\\
		\TT{V}&\mapsto 2\\
		\TT{VP}&\mapsto 3\\
		\TT{S}&\mapsto 4
	\end{align*}
on the one hand, and by
\begin{align*}
		\delta_1:\{\sqcup,\TT{NP},\TT{V}\}&\to \{0,1,2\}\\
		\sqcup&\mapsto 0\\
		\TT{NP}&\mapsto 2\\
		\TT{V}&\mapsto 1 \\	
    	\delta_2:\{\sqcup,\TT{NP},\TT{V},\TT{VP},\TT{S}\}&\to \{0,1,2,3,4\}\\
    	\sqcup&\mapsto 0\\
    	\TT{NP}&\mapsto 4\\
    	\TT{V}&\mapsto 3\\
    	\TT{VP}&\mapsto 1\\
    	\TT{S}&\mapsto 2
\end{align*}
on the other hand. Defining the step function $f: X\to \R$ as in \eqref{eq:inv_f} after choosing $(l,r)=(2,3)$ and the $c_i$-s randomly. The neural automaton consists of $n = 72$ neurons, i.e. the phase space is given by the hypercube $X = [0,1]^{72}$. Running the neural network with both encodings and computing the step function $f$ on each iteration $i=1,\dots,6$, we see in \Fig{fig:step_function} that $f$ is indeed invariant under G\"odel recoding.

The step function clearly distinguishes among different states (where here by ``different" we mean with different patterns of equality), but returns the same value for the states corresponding to the same pattern of equality, that is, states that differ on the G\"odel encodings, as desired.

\begin{figure}[H]
    \includegraphics{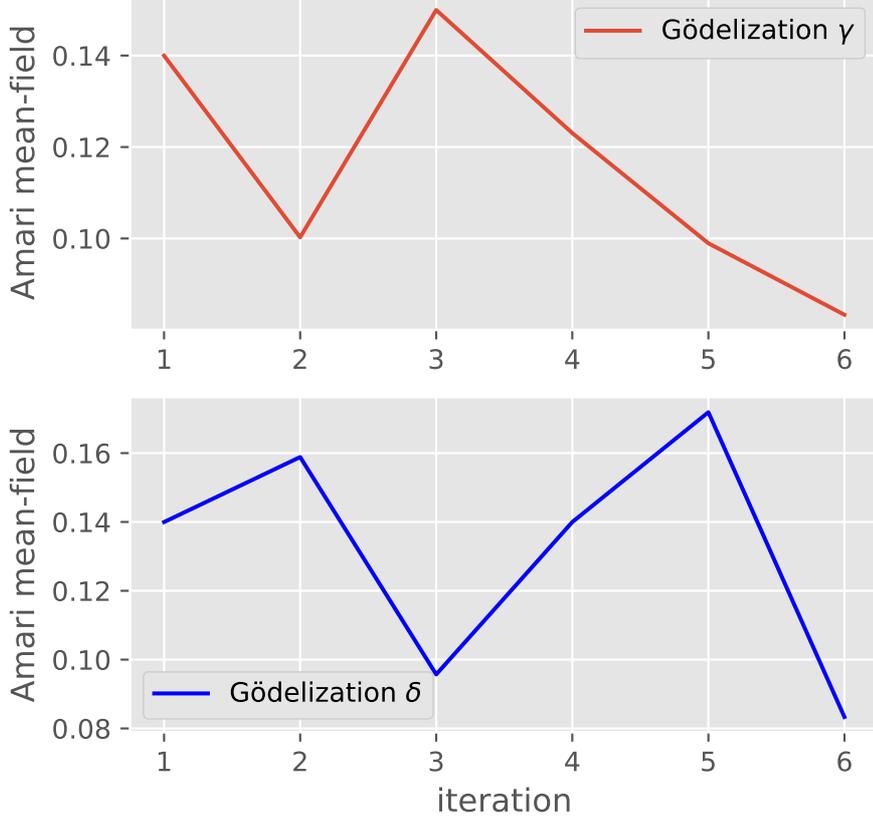}
    \caption{Amari's mean-field observable \Eq{eq:amari} of the neural automaton under two different G\"odel encodings $\gamma$ and $\delta$.}\label{fig:Amaris_observable}
\end{figure}

In contrast, if we use Amari's observable \Eq{eq:amari} for the same simulation, we obtain a very different picture, showing that this observable is not invariant under G\"odel recoding, as shown in \Fig{fig:Amaris_observable}. Obviously, this observable strongly depends on the particular G\"odel encoding we have chosen.

\section{Discussion}
\label{sec:disc}
In this study we have presented a way of finding particular macroscopic observables for nonlinear dynamical systems that are generated by G\"odel encodings of symbolic dynamical systems, such as nonlinear dynamical automata (NDA: \citet{GrabenLiebscherSaddy00, GrabenJurishEA04, GrabenGerthVasishth08, GrabenPotthast14}) and their respective neural network implementation, namely, neural automata (NA: \citet{CarmantiniEA17}). Specifically, we have investigated under which circumstances such observables could be invariant under any particular choice for the G\"odel encoding.

When mapping symbolic dynamics to a real phase space, the numbering of the symbols is usually arbitrary. Therefore, it makes sense to ask which information of the dynamics is preserved or can be recovered from what we see in phase space under the different possible options. In this direction, we have provided a complete characterisation of the strings that are and are not distinguishable after certain G\"odel encoding in terms of \emph{patterns of equality}. We have proven a partition theorem for such invariants.

In the concrete case of NA constructed as in \Citet{CarmantiniEA17}, which can emulate any Turing Machine, we have a dynamical system for a neural automaton. This system completely depends on the choice of the G\"odel numbering for the symbols on the alphabet of the NA. Based on the invariant partition mentioned before, we were able to define a macroscopic  observable that is invariant under any G\"odel recoding. In fact, by the way we define this observable, the definition is based on an invariant partition according to the length of the strings on the left and right hand side of a dotted sequence compising the machine tape of the NA. This means that each choice of the length of those strings provides a sharper invariant, making strings with different patterns of equality completely distinguishable. It is also important to mention that macroscopic observables in general are not invariant under G\"odel recoding. As a particular example, we computed the mean neural network activation originally suggested by \citet{Amari74} and later employed by \citet{CarmantiniEA17} as a modeled  ``synthetic ERP'' \citep{BarresSimonsArbib13} in neurocomputing.

In fact, any observable that is invariant under G\"odel recoding must be equally defined for points on the phase space corresponding to G\"odelizations of strings sharing the same patterns of equality. This could probably provide an important constraint in the finding of other invariant macroscopic observables.

Theoretically, one could run neural automaton under all (or many) possible G\"odel encodings and check which observables are preserved by the dynamics and which are not. This could provide important information about the performance of the neural network architecture that is intrinsic of the dynamical system, and not dependant on the choice of the numbering for the codification of the symbols. In practice, the computation of all the permutations of the alphabet grows with the factorial of the alphabet's cardinality, and the computation of invariant partitions even with powers of that number for longer strings. This, of course, would present some practical constraints for large alphabets and sharp invariant observables.

Our results could be of substantial importance for any kind of related approaches in the field of computational cognitive neurodynamics. All models that rely upon the representation of symbolic mental content by means of high-dimensional activation vectors as training patterns for (deep) neural networks \citep{Arbib95, CunBengioHinton15, HertzKroghPalmer91, Schmidhuber15}, such as vector symbolic architectures \citep{Gayler06, SchlegelNeubertProtzel21, Smolensky90, Smolensky06, Mizraji89, Mizraji20} in particular, are facing the problems of arbitrary symbolic encodings. As long as one is only interested in building inference machines for artificial intelligence, this does not really matter. However, when activation states of neural network simulations have to be correlated with real-word data from experiments in the domains of human or animal cognitive neuroscience and psychology, the given encoding may play a role. Thus, the investigation of invariant observables in regression analyses and statistical modeling becomes mandatory for avoiding possible confounds that could result from a particularly chosen encoding.

These results also have implications in Mathematical and Computational Neuroscience, where the aim is to explain by means of mathematical theories and computational modelling neurophysiological processes as observed in in-vitro and in-vivo experiments via instrumentation devices. Our results forces us to consider the possibility as to what extent (if any) that observations, which motivate the development of models in the literature (e.g Spiking models), are epiphenomenon? To conclude, we express the hope that our study paves the way towards a more a comprehensive research in computational cognitive neurodynamics, mathematical and computational neuroscience where the study of macroscopic observations and its invariant formulation can lead to interesting new insights.

\subsection{Reproducibility}

All numerical simulations that have been presented in Section \ref{sec:nlapp} may be reproduced using the code available at the Github repository \linebreak \url{https://github.com/TuringMachinegun/Turing_Neural_Networks}{}. The repository contains the code to build the architecture of a neural automaton as introduced in (\citet{CarmantiniEA17}) together with particular examples. The code that computes the invariant partitions given by equality patterns can also be found in the repository. The code allows the user to implement various observables (e.g step function, Amari's observable) in order to test further cases, exploit and further develop our framework.

\section{Acknowledgements}
SR acknowledges support from Ikerbasque (The Basque Foundation for Science), the Basque Government through the BERC 2022-2025 program and by the Ministry of Science and Innovation: BCAM Severo Ochoa accreditation CEX2021-001142-S / MICIN / AEI / 10.13039/501100011033 and through project RTI2018-093860-B-C21 funded by (AEI/FEDER, UE) and acronym MathNEURO. JUA aknowledges support from the Spanish Government, grants PID2020-117281GB-I00 and  PID2019-107444GA-I00, partly with European Regional Development Fund (ERDF), and the Basque Government, grant IT1483-22.


\appendix
\section{Proofs of lemmata and theorems}
\label{sec:apen}

\begin{proof}[of Lemma \ref{lem:iso}]
The ordering $\gamma$ itself induces the isomorphism between both graphs. Namely  let $g_{\gamma}:T_{\gamma}\to T$ be such that if $s=a_{i_1}a_{i_2}a_{i_3}\dots\in T_{\gamma}$ then $$g_{\gamma}(s)=\gamma(a_{i_1})\gamma(a_{i_2})\gamma(a_{i_3})\dots$$  which clearly belongs to $T$.

We must show that it defines a bijection between vertices and that preserves incidence.\\
It is easy to prove that it is a bijection. Namely if $w=a_{i_1}a_{i_2}\dots a_{i_n}$ and $u=b_{i_1}\dots b_{i_k}$ are any two vertices of the tree $T_\gamma$ then $g_{\gamma}(w)=g_{\gamma}(u)$ implies that both strings must have the same length, hence $n=k$. And since $\gamma(a_{i_j})=\gamma(b_{i_j})$ and $\gamma$ is a bijection, we must have $a_{i_j}=b_{i_j}$ so that $w=u$. Moreover, for any $v=l_1\dots l_n\in T$ there is $z=a_{\gamma^{-1}(l_1)}\dots a_{\gamma^{-1}(l_n)}$ which is clearly mapped to $v$ through $g_{\gamma}$.

The only thing that is left to show is that $g_{\gamma}$ preserves incidence. That is, that given $w\in\A^*$ and $a\in\A$, then $g_{\gamma}(wa)=g_{\gamma}(w)l$ for some $l\in \{0,\dots,m-1\}$. But this is also clear from the definition of $g_{\gamma}$.
\end{proof}

\begin{proof}[of Lemma \ref{lem:distance}]
Let us suppose that $d(p,q)\leq \frac{1}{m^n}$. This means that at least the first $n$ symbols in both strings are equal. Then, if $\psi$ is a G\"odel encoding defined by the asignment $\gamma:\A\to\M$ we have that
$$\psi(p)=\sum\limits_{i=1}^{\infty}\gamma(a_i)\frac{1}{m^i}=\sum\limits_{i=1}^n\gamma(a_i)\frac{1}{m^i}+\sum\limits_{i=n+1}^{\infty}\gamma(a_i)\frac{1}{m^i}.$$
Let us put $r=\sum\limits_{i=1}^n\gamma(a_i)\frac{1}{m^i}$. Now, since $\gamma(a_i)\leq m-1$
	\begin{align*}
			r&=\frac{a_1 m^{n-1}+a_2d^{n-2}+\dots a_{n-1}m+a_n}{m^n}\leq\\
			&\leq \frac{m^n-m^{n-1}+m^{n-1}-\dots-d+d-1}{m^n}=\frac{m^{n-1}}{m^n}.
	\end{align*}
So $r=\frac{k}{m^n}$ for some $k=0,\dots,m^n-1$. Since $\sum\limits_{i=n+1}^{\infty}\gamma(a_i)\frac{1}{m^i}<\frac{1}{m^{n+1}}$, we get that $\psi(p)\in\left[\frac{k}{m^n},\frac{k+1}{m^n}\right)$. Since $q$ is equal to $p$ on at least the first $n$ strings we will also have $\psi(q)=r+\sum\limits_{i=n+1}^{\infty}\gamma(b_i)\frac{1}{m^i}$ and by the same reason it will be on the same interval.

For the other implication, if we have two real numbers $\psi(p)$ and $\psi(q)$ after encoding some infinite strings $p$ and $q$, we want to show that if they are on some interval of the type $\left[\frac{k}{m^n},\frac{k+1}{m^n}\right)$, then they have the same prefix of at least length $n$. We can always write those numbers as $\psi(p)=\frac{k}{m^n}+r_1$ and  $\psi(q)=\frac{k}{m^n}+r_2$ with $r_1,r_2<\frac{1}{m^{n+1}}$. If we write the number $k$ in its $m$-adic expansion, it will be uniquely determined by $l_1,\dots,l_n\in\{0,\dots,m-1\}$, and each number $r_i$ can be writen as a series by $r_i=\sum\limits_{j=n+1}l_{j_i}\frac{1}{m^j}$, for $i=1,2$. Taking the inverse images of each $l_i$, that is $\gamma^{-1}(l_i)=a_i$ we will obtain that $p=a_1\dots a_na_{n+1}\dots$ and $q=a_1\dots a_nb_{n+1}\dots$. That is, they are at least at distance $\frac{1}{m^n}.$
\end{proof}

\begin{proof}[of Theorem \ref{thm:partition}]
Let us first assume that given $u$ there exists some $\pi$ such that its induced automorphisms maps $w$ to $u$. The condition \Eq{eq:partition_cond1} is clear, by Lemma \ref{lem:distance}. Then, we can write $u=c_1\dots c_n$. Now, we know that $g_{\pi}(w)=a_{\pi(1)}a_{\pi(2)}\dots a_{\pi(n)}=c_1\dots c_n=u$. Then
	\begin{align*}
		\{j_1,\dots,j_k\}\in P_w&\iff a_{i_{j_1}}=\dots=a_{i_{j_k}}\\
		&\iff  a_{\pi(i_{j_1})}=\dots=a_{\pi(i_{j_k})}\\
		&\iff c_{i_{j_1}}=\dots=c_{i_{j_k}}\\
		&\iff \{j_1,\dots,j_k\}\in P_u.
	\end{align*}
To show the other direction, it suffices to define $\pi$ so that $g_{\pi}(w)=u$. That is, if $u=c_1\dots c_n$, let us define $\pi(a_i)=c_i$ and let us send the $a_j\in \M$ not appearing in $w$ to the $c_j$-s not appearing in $u$ in a bijective way. This can be done, it is well defined by condition \Eq{eq:partition_cond2}, and it defines a bijection on $\M$ by construction.
\end{proof}

\begin{proof}[of Theorem \ref{thm:f_invariant}]
We have to show that equation (\ref{eq:invari}) is satisfied. Namely, we have to show that $f(\alpha^*_{\pi}(\vec{x}))=f(\vec{x})$ for each $\pi\in S_{m-1}\times S_{m-1}$ and $\vec{x}\in X$. Note that by definition, $\alpha^*_{\pi}=\rho_{\pi}$ in our case. Therefore, we must show that $f(\rho_{\pi}(\vec{x}))=f(\vec{x})$
Let $\pi=(\pi_1,\pi_2)\in S_{m-1}\times S_{m-1}$ and $(y_1,y_2)\in Y$. Then, there is some $E^{(i,j)}$ for which $(x,y)\in E^{(i,j)}$. Let us denote by $\pi'=(\pi'_1,\pi'_2)\in S_m\times S_m$ the permutation fixing $1$ and sending $\pi'(i)=\pi(i)+1$ for $i=2,\dots,m-1$ (namely, the permutation fixing the first letter and permuting the rest as $\pi_i$ permutes the $m-1$ letters for $i=1,2$ respectively).

Note that since we have enlarged our alphabet with the $\sqcup$ symbol, and since both our original encoding and the one permuted by $\pi$ send this symbol to $0$, if $(y_1,y_2)$ is decoded as $(c_1,\dots,c_l)$ and $(d_1,\dots,d_r)$, the possible $0$s appearing at the end of each encoding (indicating that the string has smaller length than $l$ and/or $r$) will remain being $0$-s, and therefore the point will not be mapped to a point encoding longer strings.

After applying $\rho_{\pi'}$, we will obtain $\rho_{\pi}((y_1,y_2))\in E^{(i',j')}$ for some $i',j'$. However, since $\rho_{\pi'}$ is defined through $g_{\pi'_1}$ and $g_{\pi'_2}$ both $E^{(i,j)}$ and $E^{(i',j')}$ belong to the same $P_{ij}$. That is, they have the same pattern of equality. Hence, by the definition of $f$ we obtain that $f(\vec{x})=f(\rho_{\pi}(\vec{x}))$, and we are done.
\end{proof}



\end{document}